\documentclass{article}

\usepackage[final]{ap_2017}
\usepackage{ap_2017}

\usepackage[utf8]{inputenc} 
\usepackage[T1]{fontenc}    
\usepackage{hyperref}       
\usepackage{url}            
\usepackage{booktabs}       
\usepackage{amsfonts}       
\usepackage{nicefrac}       
\usepackage{microtype}      
\usepackage{footnote}
\usepackage{graphics}
\usepackage{graphicx}
\makesavenoteenv{tabular}
\makesavenoteenv{table}
\usepackage{amsmath}
\usepackage{microtype}
\usepackage{array}
\usepackage[noend]{algorithm2e}
\usepackage{wrapfig}
\usepackage{color}
\usepackage{floatrow}
\usepackage{blindtext}
\usepackage{subcaption}
\usepackage{capt-of}

\newfloatcommand{capbtabbox}{table}[][\FBwidth]

\title{Slugbot: An Application of a Novel and Scalable \\ Open Domain Socialbot Framework}

\author{
	Kevin K. Bowden, Jiaqi Wu, Shereen Oraby, Amita Misra, and Marilyn Walker\\
	Natural Language and Dialogue Systems Laboratory\\
	University of California Santa Cruz\\
	\texttt{\{kkbowden, jwu64, soraby, amisra2, mawalker\}@ucsc.edu} \\
}

\begin{document}

\maketitle

\begin{abstract}
In this paper we introduce a novel, open domain socialbot for the Amazon Alexa Prize competition, aimed at carrying on friendly conversations with users on a variety of topics. We present our modular system, highlighting our different data sources and how we use the human mind as a model for data management. Additionally we build and employ natural language understanding and information retrieval tools and APIs to expand our knowledge bases. We describe our semi-structured, scalable framework for crafting topic-specific dialogue flows, and give details on our dialogue management schemes and scoring mechanisms. Finally we briefly evaluate the performance of our system and observe the challenges that an open domain socialbot faces.
\end{abstract}

\section{Introduction}
\label{sec:intro}
Personal assistant development has recently been on the rise, as systems such as Amazon Alexa continue to gain momentum in the tech industry. State-of-the-art personal assistants are primarily task-oriented and assist the user in regular household tasks such as setting timers, checking the weather, and playing music. As system capabilities increase, there is a growing interest in moving personal assistants past restricted task-oriented domains to facilitate a more friendly, human-like conversational experience with the user. 

While task-oriented domains may allow developers to focus in on expected actions or responses from users, the conversational domain naturally elicits a huge spectrum of topics and possibilities from users. To narrow down the scope of the problem, some domain specific socialbots such as Amazons MovieBot\footnote{\url{https://www.amazon.com/dp/B01MRKGF5W}} exist, but there is a strong interest in a bot that is able to engage a user about a wide variety of topics in a less restrictive setting. 

The Amazon Alexa Prize\footnote{\url{https://developer.amazon.com/alexaprize}} competition is an effort to address exactly that: advancing the state of the art in open domain socialbot interactions by providing users with a delightful, entertaining social experience. This report describes our implementation of Slugbot, our scalable, open domain socialbot that makes use of a variety of different data sources in a modular framework to provide an engaging user experience.

\section{System Design and Architecture}
\label{sec:system_arch}
\begin{figure}[h]
  \includegraphics[scale=.3]
{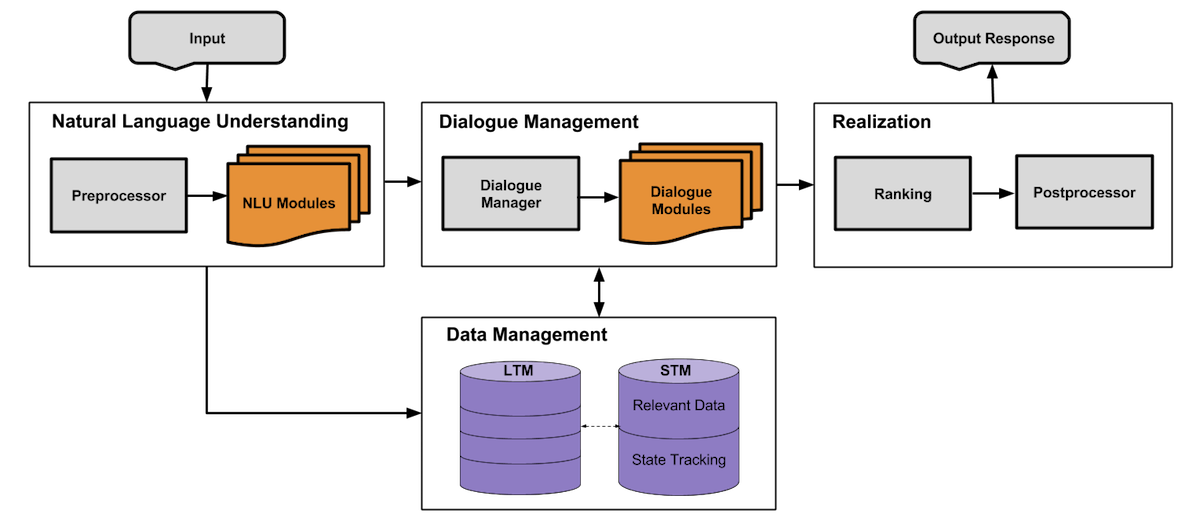}
  \centering
  \caption{Our complete system architecture.}
  \label{fig:system_arch}
\end{figure}

We begin with a look into our system design and architecture. In Sections \ref{sec:nlu} - \ref{sec:postprocess} we will describe our architecture as represented in Figure \ref{fig:system_arch}, including our natural language understanding, dialogue management, data management, and realization components. In Section \ref{sec:amazon_infrastructure} we will specifically examine our Amazon Skill infrastructure as shown in Figure \ref{fig:webapp_arch}.

\subsection{Natural Language Understanding}
\label{sec:nlu}
Our initial preprocessing stage involves analyzing the tokens interpreted by the ASR. We calculate the average hypothesis scores of the users input and prompt the user for clarification if the score is too low. Intuitively, it is better to ask for clarification rather than misinterpret the user input; however, if we consistently get a low ASR score, we are forced to estimate their utterance. To account for this, we retain all possible ASR interpretations such that we are able to better process the noise in their input.

After preprocessing the data we use our NLU engine to create a deep structure representation of the user's utterance. Our first layer of NLU relies on the Stanford CoreNLP Toolkit[10]. Our internal representation is based on the dependency parse of the respective utterance which is consolidated into a concise tree using the dependency relations. The part-of-speech (POS) and sentiment score from CoreNLP are also encoded into this structure. We do coreference resolution by mapping the coreference tags returned by CoreNLP to the data stored within our system. Finally, we use a homegrown named entity recognizer and topic classifier in addition to the NPS[6] dialogue act classifier to make our internal representation as robust as possible.

\subsection{Data Management}
\label{sec:data}
Our system uses the human mind as a model for managing memory. To our knowledge, we are the first to attempt managing system memory within a dialogue system using this model. Specifically, we differentiate between long term memory (LTM) and short term memory (STM). LTM and STM communicate with each other to exchange data: the data which is out-of-date should transfer into the LTM while the conversationally relevant data should be made available in the STM. Both memory clusters are comprised of a network of memory nodes which each have their own responsibilities. 

Our STM cluster is responsible for managing our system state and handling data which is localized to specific functionality. This local memory helps us to improve the efficiency of data access and reduce the workload of remote databases.

Our LTM cluster is responsible for managing our corpora and other large datasets. We initially utilized DynamoDB, but it was not convenient for real time fuzzy queries and frequent updates. We instead utilized the more flexible RDS Relational Database to store real-time search data in addition to other scraped data relevant to our various modules. Finally, it is within this cluster where we will perform our reinforcement learning.


\subsection{Dialogue Manager}
\label{sec:agent_core}

\begin{table*}[ht!]
\begin{small}
\begin{center}
\begin{tabular}
{p{2cm}|p{1cm}|p{10cm}}
\toprule
\bf Module & \bf Section & \bf Description \\
\midrule
Base Responses & \ref{sec:agent_core} & State-specific responses like handling repeat requests or prompting with a menu\\ \midrule
Opinions & \ref{sec:reactive} & Solicit, provide, and justify opinions about detected entities \\ \midrule
Question Answering & \ref{sec:reactive} & Question answering modules including ELIZA, Evi, Wikipedia, and DuckDuckGo \\ \midrule
Retrieval & \ref{sec:reactive} & Elasticsearch index used to retrieve appropriate responses \\ \midrule
Out-of-Domain & \ref{sec:reactive} & Out-of-Domain responses to sustain the conversation if there are no other good options \\ \midrule
Storytelling & \ref{sec:active} & Tell the user a story and answer questions based on a corpus of personal narratives \\ \midrule
Games & \ref{sec:active} & Games such as \textit{Jeopardy}, \textit{Fast Money}, the City Name game, \textit{Nim}, and story adventures \\ \midrule
Surveys & \ref{sec:active} & Generally pop-culture themed surveys consisting of around 5 questions before giving results \\ \midrule
Interactive Sequences & \ref{sec:active} & Short 2 turn sequences which can be recursively triggered repeatedly, including riddles and would you rather questions \\ \midrule
Recursive & \ref{sec:active} & Facts or trivia triggered recursively until user wants to change topics \\ \midrule
Flow Manager & \ref{sec:flowman} & Managing dialogue flows about a variety of topics (currently, 31) as a way to rapidly increase popular topic coverage \\
\bottomrule         
 \end{tabular}
 \caption{Summarized list of different modules.}
\label{table:modules}
\end{center}
\end{small}
\end{table*}

The dialogue manager is tasked with handling the most basic functionality, such as detecting repeat requests, stop requests, and prompting the user with a menu of topics to help transition into domains we are more familiar with. We maximize the number of unique experiences within a single conversation by prioritizing unexplored topics. We found that having a menu made an appreciable difference on the quality of the conversation.

The dialogue manager also sets system expectations. Specifically, we indicate the data we are expecting from the user, representing the preconditions which must be satisfied in order for a particular action to be valid. Our expectations can be satisfied by observing attributes from the systems state. We can define our expectations using a variety of state variables including direct keyword matching, specific utterance attributes such as dialogue act or sentiment score, or even be contingent on the result returned by a function. 

\subsection{Dialogue Modules}
\label{supplementary_cores}

Table~\ref{table:modules} lists the dialogue modules implemented in our system, with descriptions of each module in the following sections. Our dialogue manager maintains control over the dialogue modules, allowing us to easily change the type of responses which we pool. We feature primarily two module classes, mixed initiative in the case where our agent does not have control of the conversation, and system initiative when the agent is driving the conversation.

\subsubsection{Mixed Initiative Modules}
\label{sec:reactive}


Mixed initiative modules are designed to solicit an even exchange in content per turn from both the user and the agent. More specifically, this is the case in which the agent is not strictly in control of the conversation and in fact, our content is likely to be a reaction to the users initiative. Such user initiative could be asking a question, soliciting an opinion, or having general chit-chat outside of a controlled dialogue flow. 

\textbf{Opinions:} Here we can learn more about the user by asking for their name or soliciting their opinion of a contextually relevant entity. Naturally if we can solicit user opinions it's important that we are able to provide and justify our own opinions. To accomplish this, we load our agents profile with opinions about various entities and abstract concepts. The first time a user engages with the system we randomly select opinions to include in the agents profile, allowing us to take the preliminary steps towards giving the agent a unique and identifiable personality. Our opinion dataset is extracted from online reviews of movies, video games, and books. We have also handcrafted data points which would give us good general coverage of conceptual opinions based on popular topics which would be difficult to extract from any data source such as \textit{What is your favorite color?} Table \ref{table:opinions_ex} demonstrates how we are able to leverage our structured opinion data to answer solicitations from the user.

\begin{wraptable}{L}{.5\textwidth}
\begin{small}
\centering
\begin{tabular}
{p{0.9cm}|p{5.1cm}}
\toprule
User & {What's your favorite film?} \\
Agent & {I loved The Terminator.} \\
User & {Why do you like The Terminator?} \\
Agent & {I think the Terminator is action packed and well cast.}\\
User & {What's your favorite color?} \\
Agent & {I really like purple, I think it's beautiful.}\\
User & {What do you think of popcorn?} \\
Agent & {I hate popcorn, it's just so greasy.}\\
\bottomrule         
 \end{tabular}
 \caption{Sample usages of our opinion data.}
 \label{table:opinions_ex}
\end{small}
\end{wraptable}

\textbf{Question Answering:} Our question answering mechanism is a three step inspection of the query. First, if there are not enough content words to detect the intent of the question, we probe for more information using a modified version of ELIZA[12]; we found that users generally reacted poorly to some of the more intimate prompts. If the system has the initiative, such as when we are telling a story, we assume the associated module will be able to answer questions using the module's structured data. If neither of these conditions are true or we don't yet have an answer, we perform coreference disambiguation and query our search engine. Our search engine consists of three search components which are queried in the following order: Evi \footnote{\url{https://www.evi.com/}}, Wikipedia \footnote{\url{https://www.wikipedia.org/}}, and finally DuckDuckGo \footnote{\url{https://duckduckgo.com/}}. If we've still not returned an answer we will inform the user that at this time we are not able to answer the question. We noticed through anonymous user feedback that in general users were okay with our system not being able to answer questions as long as the agent made it clear that it wasn't sure of an answer. An example demonstrating this module is shown in Table \ref{table:qa_ex}.

\begin{wraptable}{L}{0.5\textwidth}
\begin{small}
\begin{tabular}
{p{0.9cm}|p{5.1cm}}
\toprule
User & {What is the capitol city of Mexico.}\\
Agent & {The capitol city of Mexico is Mexico City. (search)} \\
User & {What is it's population?}\\
Agent & {The population of Mexico City is 8.8 million. (disambiguation + search)} \\
User & {Okay, how is it that you are smart?}\\
Agent & {Why do you think I am smart? (ELIZA)}\\
User & {Just a guess. Tell me a story.}\\
Agent & {Did I ever tell you one time my pet Moses really scared me? (enter storytelling)}\\
User & {No, what kind of pet is it?}\\
Agent & {Moses is a chinchilla. (storytelling answer)}\\
\bottomrule               
 \end{tabular}
\end{small}
\caption{Sample question answering}\label{table:qa_ex}
\end{wraptable}

\textbf{Retrieval:} There are several existing retrieval based chatbots which operate on large existing corpora such as Twitter[11, 8], the Open Subtitles corpus[5], or movie scripts[3,2]. While this approach works well for specific user utterances, it generally performs worse than other dialogue management methods[8]. Therefore our system uses retrieved responses sparingly. Our retrieval module is constructed using an Elasticsearch\footnote{\url{https://github.com/elastic/elasticsearch}} based index. While large datasets for retrieval systems do exists, the utterances are not tailored towards a two-party open domain conversation, motivating us to crowd source a high quality corpus of turns over a variety of topics using qualified workers on Mechanical Turk. From this data collection we gathered 17,169 topic annotated turns. In addition to our crowd sourced data collection we further supplemented our index by hand crafting responses which would have only been captured by observing them in our user trials as discussed further in Section \ref{sec:discussion}.

\textbf{Out-of-Domain:} Using a combination of these modules we also designed our out-of-domain responses for when we aren't confident in any specific answer. For example, if we are able to detect an entity but don't have a direct follow-up utterance, we can either given an opinion, ask the user for more information, verify the entity by asking about one of it's related synonyms, or use our search engine to try and retrieve some possible summary-based response. If we are still unable to get a suitable candidate we will transition to a new topic with the help of hedge phrases such as \textit{Moving on,} or \textit{Anyways}. We rely on our scoring, discussed in Section \ref{sec:scoring}, to pick a new topic which is as relevant as possible to the users most recent utterance.

\subsubsection{System Initiative Modules}
\label{sec:active}

Our system initiative modules gain control of the conversation by driving the content or doing an activity with the user. These modules are designed to encourage extended multi-turn interactions by routing the user back to the module when the conversation starts to stray. 

\begin{wraptable}[15]{L}{0.5\textwidth}
\centering
\begin{small}
\begin{tabular}
{p{0.9cm}|p{5.1cm}}
\toprule
Agent & {Do you want to hear some science facts?} \\
User & {Yes} \\
Agent & {Did you know that At over 2000 kilometers long, The Great Barrier Reef is the largest living structure on Earth. Want to hear another?} \\
User & {Sure why not} \\
Agent & {How about this one. The average human body carries ten times more bacterial cells than human cells. Want to hear more?} \\
\bottomrule               
 \end{tabular}
\end{small}
  \caption{Sample recursion for the science topic.\\}\label{table:recursive_ex}
\end{wraptable}

\textbf{Stories:} Stories are often told conversationally and in a two party setting it is natural that one speaker will assume the responsibility of content provider while the other will assume the role of recipient. This can lead to an entertaining experience for the recipient while also minimizing the conversation's reliance on their content. Therefore, if we are able to replicate this environment we hope to create an engaging multi-turn experience. From a large corpus of personal blogs[9] we build a small dataset of around 40 narratives which are interesting and don't contain inappropriate content. On average the narratives ranged between 10-20 sentences, where our system would step through 1-3 sentences per turn until the story had been completed. Additionally, we hand annotated each story with question answer pairings such that we were able to give further details about each story. We found that users preferred hearing about stories which were generally more positive in tone and despite limited QA functionality, users tended to enjoy stepping through the story slowly rather than having the entire story in one long utterance. 

\begin{wraptable}[19]{L}{0.5\textwidth}
\begin{small}
\begin{tabular}
{p{0.9cm}|p{5.1cm}}
\toprule
Agent & {How about I ask you some would you rather questions?}\\
User & {Okay} \\
Agent & {Would you rather win the Nobel Peace Prize or 5 million dollars?}\\
User & {Oh I don't know that's a tough question, I guess I would want to win a Nobel Peace Prize.} \\
Agent & {Interesting, I would choose the first option too. I would rather win the Nobel Prize because it would mean that I have done something significant instead of just being handed some money for no good reason. Anybody can win 5 million dollars but not everyone can win the Nobel Prize. Want to hear another?}\\
\bottomrule               
 \end{tabular}
\end{small}
\caption{Sample would you rather sequence.}\label{table:sequence_ex}
\end{wraptable}

\textbf{Games:} We support a variety of different games which the user can play including \textit{Jeopardy}, \textit{Fast Money}, the \textit{City Name} game, and \textit{Nim}. These games are built on top of data dumps which were processed to make them more appropriate for the spoken domain. Additionally, the user is able to play a \textit{text adventure}, a game in which the agent and the user build a story together. Text adventures seem like a highly effective means of both entertaining the user and soliciting real dialogue with the agent. We use 10 different prompts which have been extracted from the internet[1]. The user is also able to take various quizzes which often revolve around pop culture topics such as {\it "Which Harry Potter house do you belong to?"} and subsequently include a set of questions which inevitably resolve in the system classifying the user. Our survey data was extracted from popular quizzing sites and then further refined based on the topics which users found interesting. We found empirically that quizzes about pop culture entities and personality type were the most popular. Despite some difficulties working with the ASR, in generally these activities were effective at engaging the users and creating extended multi-turn conversations.

\textbf{Recursive:} Finally, we have a set of modules which acts recursively. Here we can inform the user of various headlines from new sources, or give the user facts about a topic of their choice. We are able to recurse over this functionality by simply continuously giving them information until they explicitly transition out of the recursion. We have included an example of this in Table \ref{table:recursive_ex}. We can also create recursive 2-turn sequences by asking the user a sequence of riddles, or asking \textit{would you rather} questions. Both of these cases allow the user and agent to converse for a couple of turns briefly about the sequence before recursing, an example of this can be seen in Table \ref{table:sequence_ex}. These recursive loops are highly effective at keeping the user engaged with the agent in a multi-turn context without having to worry about a complex dialogue flow.

\subsubsection{Flow Manager}

One of the primary modules in our system, the flow manager, is responsible for managing the flow of dialogue related to a given topic or utility. A flow, as seen in Figure 2, is organized in a graph structure where each node has specific preconditions, postconditions, and actions which work together to sustain a natural dialogue about any particular subject. While flows in general are meant to provide coverage of some specified root topic, like Books, it's important to note that many flows contain multiple subroots, such as "favorite genre", "book trivia", or "bestsellers". The user can directly trigger a flow about a given topic by using related keywords, or by expressing interest in the flow if the prompt is selected by the system when propositioning topics.

\label{sec:flowman}
\begin{figure}
\centering
\begin{floatrow}
\capbtabbox{%
  \includegraphics[scale=.4]{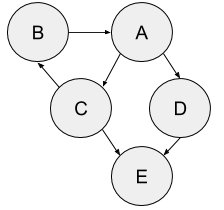}
  \label{fig:sample_flow}
}
{%
  \captionof{figure}{Sample flow.}%
}
\capbtabbox{%
  \begin{tabular}
{@{}p{0.1cm}|p{0.9cm}|p{4cm}@{}}
\toprule
1 & User & {Precondition A} \\
2 & Agent & {Action A, Postcondition A} \\
3 & User & {Precondition C} \\
4 & Agent & {Action C, Postcondition C} \\
5 & User & {Precondition B} \\
6 & Agent & {Action B, Postcondition B} \\
7 & User & {No Precondition} \\
8 & Agent & {Exit Flow} \\
\bottomrule         
 \end{tabular}
}{%
  \caption{A sample conversation.}%
}
\capbtabbox{
\begin{tabular}
{@{}p{0.1cm}|p{2cm}@{}}
\toprule
1 & {A, .... ,Z} \\
 & \\
3 & {C, D} \\
 & \\
5 & {B, E} \\
 & \\
7 & {A} \\
 &  \\
\bottomrule         
 \end{tabular}
}{%
  \caption{Expecting.}%
}
\end{floatrow}
\end{figure}

\begin{wraptable}[20]{L}{0.71\textwidth}
\begin{small}
\begin{tabular}
{l|c|c}
\toprule
Flow & Samples & Score\\
\hline
Astronomy & 20(50) & 3.48\\
Board Games & 17(43) & 3.94\\
Books & 20(68) & 3.0\\
Box Office & 18(43) & 3.31\\
Comic Books & 5(51) & 3.6\\
Dinosaur & 11(44) & 3.77\\
Favorite Food & 32(65) & 3.31\\
Fun Facts & 17(45) & 4.15\\
Gossip & 25(60) & 3.6\\
Health & 7(55) & 3.43\\
History & 26(52) & 3.46\\
Hobbies & 21(51) & 3.88\\
Holidays & 11(51) & 3.23\\
Horoscope & 34(62) & 3.6\\
Joke & 54(96) & 3.94\\
Monsters & 46(55) & 3.14\\
\bottomrule 
 \end{tabular}
 \begin{tabular}
{l|c|c}
\toprule
Flow & Samples & Score\\
\hline
Movie & 41(168) & 3.45\\
Music & 66(139 & 3.36\\
News & 27(59) & 3.28\\
Poem & 19(40) & 3.29\\
Quote & 29(59) & 3.48\\
Recipe & 19(42) & 3.66\\
Science & 24(51) & 3.29\\
Shopping & 13(46) & 3.08\\
Sports & 58(149) & 3.2\\
Style & 7(33) & 3.14\\
Technology & 170(210) & 3.25\\
Travel & 17(50) & 3.35\\
TV & 10(48) & 3.1\\
Video Games & 16(69) & 3.75\\
Weather & 29(73) & 3.6\\
& & \\
\bottomrule         
 \end{tabular}
 \centering
 \caption{Our current flows, here Samples are \#utilized(\#prompted).}
 \label{table:flows}
\end{small}
\end{wraptable}

Our preconditions are represented by the expected data discussed in Section \ref{sec:agent_core}. Postconditions can represent a variety of desirable effects which only occur after the response has been realized by the dialogue manager. It is within these postconditions that we can indicate calls to external functions or update specific state variables. Finally, there are actions which also occur. These actions can modify a candidate utterance or delegate responsibility for response curation to a different module.

Flows represent a high level abstraction of our entire systems functionality, allowing a new designer to rapidly add content to the system without needing to familiarize themselves with the underlying architecture. As seen in Table \ref{table:flows}, our system currently supports 31 flows covering a diverse range of topics. In Table \ref{table:flows} we have also included the number of samples in which the user actually utilizes a given flow (>2 turns) vs. the number of times the user was prompted to start the flow in parenthesis, along with the average score a conversation will receive when a flow is utilized. Since these flows represent a high level abstractions of the entire system, we found that reusing successful modules is an effective means of bootstrapping flows with minimal effort. Specifically, most flows have some recursive trivia based prompt in their list of subroots. We also found that generically discussing user preferences and utilizing search methods increased the breadth of a flow, while a combination of all methods could increase the depth of a flow.

\subsection{Scoring}
\label{sec:scoring}

Once we've established a pool of responses we rerank them to find a response we are most confident in. Specifically, each response has a confidence score which ranges from 0 to 1. Each response is assigned a base confidence when it is added to the pool - this base confidence will vary significantly based on the origin of the response and the users utterance. If the user says \textit{I like video games} for example, the \textit{video game} conversation starter will have a base confidence of 1. If the user said \textit{I like dogs}, the \textit{video game} prompt would be initialized with a confidence of .6, indicating it as a valid topic starter but only if we have nothing more relevant to say. At this stage our sensitive content filter will invalidate any response with explicit content and detect if a priority response has been triggered. Priority responses are valid regardless of our current state and indicate responses which are to be uncontested - such as repeat requests or stop session markers. Finally, for all other responses we update their confidence using Equation \ref{eq:scoring}. We attempt to increase our confidence in the response by looking for contextually relevant content and inspecting the current system state. Our context score is calculated based on overlapping content words and entities and our inspection of our system's system.

Equation \ref{eq:loss} represents how we penalize a given response. A response is in a state of incoherence if it does not belong to the current system initiative module. For example, if we are playing a specific game, and a response stems from anywhere besides that game it would create incoherence within the conversation. In order to maintain module coherence we apply an empirically derived .15 penalty to these responses. While we leverage the state tracking done by our short term memory to avoid repetitious utterances, some general prompting phrases such as {\it "would you like to play a game?"} are still valid despite being already said. In order to increase the diversity of a user's experience without limiting the variety in our response pool we apply a .05 penalty to prompts which we have already been explored. Furthermore, we noticed from own experimentation that long utterances from mixed initiative modules tended to be received poorly. An example of this includes long news headlines and overly verbose indexed responses, where we have limited control over the phrasal timing. We therefore applied a length based penalty to these utterances.

\begin{equation} 
\label{eq:scoring}
\textbf{def score:  } r_{i}.confidence = min(max(context(r_{i}), r_{i}.confidence) - loss(r_{i}), 1)
\end{equation}

\begin{equation} \label{eq:loss}
\textbf{def loss:  } r_{i}.confidence = incoherence(r_{i}) + repeat(r{i}) + sentLen(r_{i})
\end{equation}

In the case that more than one response shares the highest confidence score, we resolve ties by randomly choosing among the candidates. It should be noted that in most cases candidates which are tied for the maximum score all tend to realize the same intention.

\subsection{Postprocessing}
\label{sec:postprocess}
Once we have selected a response we do some slight surface modifications to the text. We use a statistical natural language generation engine [4] to add variation to the start of phrases when necessary. Moreover, at this stage we apply any SSML tags which are encoded within our response. While SSML certainly holds potential to improve the user experience, when realized most of the tags other than \textit{pauses} made the utterance sound odd and detracted from the conversation. 

\subsection{Web Application Architecture}
\label{sec:amazon_infrastructure}



\begin{figure}[h]
  \includegraphics[scale=.25]{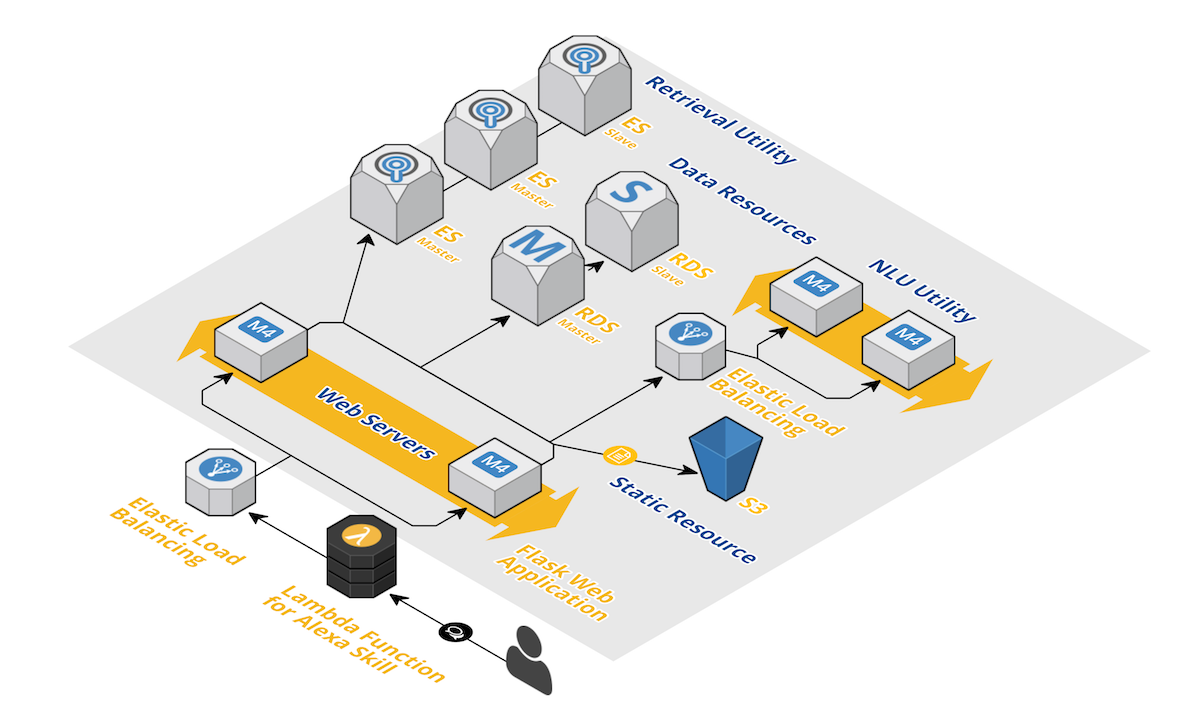}
  \centering
  \caption{Our Web Application Architecture.}
  \label{fig:webapp_arch}
\end{figure}

Figure ~\ref{fig:webapp_arch} depicts the architecture of our web server in details.

\textbf{Flask Web Application:} We apply the Flask web application structure to facilitate communication between our framework and AWS Lambda. We consider exception handling an important part of our system design. Not only do we keep the daily user interaction logs in EC2 for manual inspection, but we also found the CloudWatch Alarms useful for error notifications.

\textbf{Elastic Load Balancer:} Elastic Load Balancer was easier for us to incorporate in our existing framework than Elastic Beanstalk, even though Elastic Beanstalk is convenient for deployment. To validate our load balancing, we use the Locust\footnote{http://locust.io/} load testing framework to expose potential hazards.

\textbf{Natural Language Understanding (NLU):} Our NLU utilities, specifically coreNLP, runs on its own server and could represent a performance bottleneck especially in the case of long utterances. Therefore we run these NLU utilities on a separate EC2 instance which communicates with the web server.

\textbf{Data Resources:} We use a vanilla RDS Relational Database (Aurora MySQL) with only one replication to store our data resources. Originally we used NoSQL DyanmoDB, however we found that updating and searching in real time was not practical. 

\textbf{Retieval Utility}: We are running an Elasticsearch server on a separate EC2 instance to support our retrieval based modules and future reinforcement learning in real time. 

\section{Evaluation}
\label{sec:evaluation}

\begin{wraptable}[16]{L}{0.35\textwidth}
\begin{small}
\centering
\begin{tabular}
{l|c|c}
\toprule
Signature & \# Turns & Score\\
\hline
Storytelling & 7.78 & 3.31\\
Recursive & 5.08 & 4.0\\
\textbf{Games:} &  &\\
CYOA & 6.49 &3.93\\
City Names & 6.45  &3.73\\
Fast Money & 21.83 &3.96\\
Jeopardy & 17.64 &3.66\\
Number Game & 5.75 &3.83\\
Survey & 4.10 & 3.70\\
\textbf{Sequences:} &  &\\
Riddles & 6.72 &3.33\\
WYR & 6.33 &3.71\\
\bottomrule 
 \end{tabular}
 \caption{System Initiative Modules.}
 \label{table:active_topics}
\end{small}
\end{wraptable}

We claim that an entertaining socialbot which can engage the user in a variety of activities other than pure conversation will lead to a better user experience. In Table \ref{table:active_topics}, we look at the different modules discussed in Section \ref{sec:active}. Here, we see the number of turns the user interacted with the specified module and the average score of a conversation in which the interaction happened. We note that the number of turns here do not account for the time in which the user is in a modules menu; for example, while they are picking a game to play or a survey to do. We have also removed from consideration the case in which the user does not actually engage with the activity, for example when they enter a menu but change topics. 

We can see from these results that our prediction is validated - conversations in which the user participated in an activity were on average rated higher than our teams overall average (3.1 at the end of the semifinal user feedback period). Furthermore, we can see that these activities were very effective at keeping the user engaged in a multi-turn interaction. In this report we only count a turn for each user's turn.

In Table \ref{table:recursive_topics} we look more closely at the topics for which the user was able to select when using the recursive module. We can see that all of these topics seem to effectively sustain multi-turn interactions. As seen in Table \ref{table:recursive_ex}, these interactions are very straight forward and only require an small dataset of topic related utterances to function. Moreover, we see that the conversations which contain these recursive modules also tend to have higher scores on average. We notice specifically the history related topics, which both have an average score of 5.0, this indicates to us that users interested in hearing about history were entirely satisfied by a decently sized dataset of history facts. \\

\begin{wraptable}[24]{L}{0.38\textwidth}
\begin{small}
\centering
\begin{tabular}
{l|c|c}
\toprule
Topic & \# Turns & Score\\
\hline
Ancient Wonders & 12.0 & 5.0\\
Astronomy & 7.38 & 3.69\\
Box Office & 3.0 & 3.40 \\
Comic Books & 3.33 & 4.67\\
Dinosaurs & 3.5 & 3.50\\
Fashion & 4.07 & 3.37\\
Fun Facts & 5.83 & 4.17\\
Gossip & 5.5 & 4.17\\
History Eras & 4.0 & 5.0\\
History Facts & 6.0 & 5.0\\
Jokes & 4.87 & 4.17\\
Modern Wonders & 0.0 & 0.0\\
Music & 6.36 & 3.95\\
Natural Wonders & 5.0 & 4.0\\
Poems & 3.0 & 5.0\\
Tech Products & 5.6 & 3.5\\
Quotes & 4.33 & 3.17\\
Science & 4.55 & 3.61\\
Self Driving Cars & 5.0 & 3.60\\
Sports & 4.57 & 3.39\\
Virtual Reality & 4.31 & 3.72\\
\bottomrule 
 \end{tabular}
 \caption{Different recursive topics.}
 \label{table:recursive_topics}
\end{small}
\end{wraptable}

\section{Discussion}
\label{sec:discussion}
We noticed specific challenges which using the Amazon Echo device as an interface for a socialbot. Most glaringly, users desire for the bot to perfectly assume the responsibility of a personal assistant in addition to social capabilities. Users want to adjust the volume of the device, play music, or perform other standard Echo skills. If this kind of full pipeline integration was possible, we believe it would have greatly improved the overall user experience. Variable lag times between responses due to unreliable internet connectivity also makes naive search less practical, and often testers would interact with the skill while distracted, leading to noisy input.

The largest difficulty in using the device was ASR misinterpretation. While this is surely always a source of frustration in spoken dialogue systems, we found that in an activity based environment where a user is penalized for wrong answers, or trying to answer survey questions, the frustration is magnified. We tried resolving this by inspecting all of the possible ASR interpretations, however this did not yield perfect accuracy. As a result we had to design our data in such a way to avoid these words and phrases as much as possible. We also modified the standard Amazon StopIntent and CancelIntent to handle additional cases encountered in extended conversations. As the conversation progressed, it became more probable for the device to misinterpret short user input as a stop intent, prematurely ending the conversation. 


From some of our anonymous feedback we were able to explore user preferences within the system. We noticed that explicitly reaffirming the same entity mentioned by the user made the user feel more engaged in the conversation and that being able to provide evaluative comments was received favorably. It was also clear that we should have a sentiment filter for news headlines and story content; users expressed disinterest in hearing about this negative content. Finally, we found that users placed a higher value on precision over recall in that they preferred we state when we aren't sure what to say next instead of just giving a poor answer.

In general, users seemed to enjoy the various activities and they certainly held the users engagement as they tended to result in long multi-turn interactions. It could be argued that this functionality doesn't explicitly represent a socialbot but more of an entertainment bot, but in fact in many real world social environments, games and stories are often used to break the ice between strangers, so it follows that this functionality is also appropriate when engineering socialbots. This idea is echoed by recent work[7] which confirms that in a semi-casual multi-party social environment much of a conversation consists of such activities. 

\section{Future Work and Conclusion}
\label{sec:future_Work}
In this paper, we have presented novel first steps towards a scalable open domain socialbot which uses the human mind as a model for data management. In addition to many interesting approaches to encourage user engagement, we have presented dialogue flows, a scalable design which makes it possible to easily extend our system to new domains while also dynamically altering system expectations. 

Clearly, however, this problem remains unsolved and there is much future work ahead of us. It was clear that making evaluative statements regarding various entities was well received by the user, however our methods for curating these opinions are inefficient and future work will have us exploring more novel approaches to automatic opinion generation. While we are confident that our scoring algorithm is reasonably successful at picking appropriate responses, a logical next step is to see how augmenting our algorithm to include reinforcement learning will effect the output. While our set of flows is general enough to have wide topic coverage, it is clearly an intractable task to make a flows with high levels of granularity for every topic in the open domain. Therefore it follows that finding a method to automatically generate flows from user input, and moreover automatically generating expectations which can be dynamically utilized throughout the system, is ideal. 

Finally, additional future work would be to test the system in a more realistic user environment and further expand on the functionality which is contingent on the previous user sessions. By exploring this functionality we predict that we can greatly increase the feeling of personalization with the bot as we are able to use entrainment to better sync our personality with the user and build on their previous interactions for a more unique experience.  

\section*{References}
[1] Text Adventures. Text Adventures. \url{http://editthis.info/create_your_own_story/}.\\

[2] David Ameixa, Luisa Coheur, Pedro Fialho, and Paulo Quaresma. Luke, I am Your Father:
Dealing with Out-of-Domain Requests by Using Movies Subtitles, pages 13–21. Springer
International Publishing, Cham, 2014.\\

[3] Rafael E. Banchs and Haizhou Li. Iris: A chat-oriented dialogue system based on the vector
space model. In Proceedings of the ACL 2012 System Demonstrations, ACL ’12, pages 37–42,
Stroudsburg, PA, USA, 2012. Association for Computational Linguistics.\\

[4] Kevin K. Bowden, Grace I. Lin, Lena I. Reed, Jean E. Fox Tree, and Marilyn A. Walker. M2d:
Monolog to dialog generation for conversational story telling. In Frank Nack and Andrew S.
Gordon, editors, Interactive Storytelling, volume 10045 of Lecture Notes in Computer Science,
pages 12–24. Springer International Publishing, 2016.\\

[5] Guillaume Dubuisson Duplessis, Vincent Letard, Anne-Laure Ligozat, and Sophie Rosset.
Purely corpus-based automatic conversation authoring. In LREC, 2016.\\

[6] Eric N. Forsythand and Craig H. Martell. Lexical and discourse analysis of online chat dialog.
In Proceedings of the International Conference on Semantic Computing, ICSC ’07, pages 19–26,
Washington, DC, USA, 2007. IEEE Computer Society.\\

[7] Emer Gilmartin, Benjamin R. Cowan, Carl Vogel, and Nick Campbell. Chunks in multiparty
conversation - building blocks for extended social talk. 2017.\\

[8] Ryuichiro Higashinaka, Kenji Imamura, Toyomi Meguro, Chiaki Miyazaki, Nozomi Kobayashi,
Hiroaki Sugiyama, Toru Hirano, Toshiro Makino, and Yoshihiro Matsuo. Towards an opendomain
conversational system fully based on natural language processing. In Jan Hajic and
Junichi Tsujii, editors, COLING, pages 928–939. ACL, 2014.\\

[9] Stephanie M. Lukin, Kevin Bowden, Casey Barackman, and Marilyn A. Walker. A corpus of
personal narratives and their story intention graphs. In Proceedings of the 10th International
Conference on Language Resources and Evaluation (LREC), 2016.\\

[10] Christopher D. Manning, Mihai Surdeanu, John Bauer, Jenny Finkel, Steven J. Bethard, and
David McClosky. The Stanford CoreNLP natural language processing toolkit. In Association
for Computational Linguistics (ACL) System Demonstrations, pages 55–60, 2014.\\

[11] Lasguido Nio, Sakriani Sakti, Graham Neubig, Tomoki Toda, Mirna Adriani, and Satoshi
Nakamura. Developing Non-goal Dialog System Based on Examples of Drama Television, pages
355–361. Springer New York, New York, NY, 2014.\\

[12] Joseph Weizenbaum. Eliza—a computer program for the study of natural language communication
between man and machine. Communications of the ACM, 9(1):36–45, 1966.\\


\end{document}